# Serial fusion of multi-modal biometric systems

Gian Luca Marcialis, Paolo Mastinu, and Fabio Roli
*University of Cagliari – Department of Electrical and Electronic Engineering*
*Piazza d'Armi – I-09123 Cagliari (Italy)*
{marcialis, roli}@diee.unica.it; paolomastinu@gmail.com

**Abstract**

*Serial, or sequential, fusion of multiple biometric matchers has been not thoroughly investigated so far. However, this approach exhibits some advantages with respect to the widely adopted parallel approaches. In this paper, we propose a novel theoretical framework for the assessment of performance of such systems, based on a previous work of the authors. Benefits in terms of performance are theoretically evaluated, as well as estimation errors in the model parameters computation. Model is analyzed from the viewpoint of its pros and cons, by mean of preliminary experiments performed on NIST Biometric Score Set 1.*

## 1. Introduction

Sequential, or serial, fusion of multiple matchers is a good trade-off between the widely adopted parallel fusion and the use of a mono-modal verification system [1]. In serial fusion, only one matcher is involved and other ones are requested, sequentially, according to a certain criterion. These systems exhibit the same macro-characteristics of a parallel one: multiple biometrics, multiple hardware, etc. Therefore, their acceptability is due to two additional advantages: (a) the majority of genuine users should be accepted by using only one biometric, that is, the first one in the processing chain (this can be particularly true if some partitioning of users is possible [9]); (b) all available biometrics should be required to unauthorized users.

A lot of attention has been paid to parallel fusion at feature-level, sensor-level and decision-level [1]. A few of works dealt so far with "serial" fusion. For example, in the case of a bi-modal biometric system, where matchers are combined serially, the second biometric should be requested if making a decision about the person's identity is not possible by the first one [2-4].

State of the art about serial fusion of multimodal biometric systems is basically given by Refs. [2-4, 10-11]. Refs. [2, 4] use the Wald test, or Sequential Probability Ratio Test (SPRT), for deciding the subject classification or requiring further biometrics. This method generalizes the Neyman-Pearson approach [5]. In [3], we proposed an alternative approach, which meets (a) and (b) requirements as well. This approach can be followed when estimation of multi-modal likelihood ratio is difficult due to lack of samples. In fact, whilst [2, 4] require the estimation of individual probabilities of genuine users and impostors given the matching scores, [3] is only based on ROC curves of individual matchers. Refs. [10-11] uses basically the same concept of "uncertainty region", then investigated in [3].

The framework proposed in [3] is affected by some limitations: only the two matcher case is considered, and, secondly, it does not take into account for estimation errors affecting individual ROC curves. Hower, Ref. [3] clearly proved and showed analytical relationships between overall system performance and individual systems, especially for designing the most appropriate processing chain. thus suggesting some guidelines for designers.

In this paper, the first contribution to the state of the art is that methodology reported in [3] is generalized to more than two matchers. The rationale behind our proposal is to obtain a simple and effective methodology for serial fusion of multiple matchers, whose behaviour can be predicted analytically from individual ROC curves. Such a framework can be very useful for the designer of multi-modal biometric systems, in order to "set" the best sequence of matchers without dealing with the combinatorial explosion of experiments usually needed for this assessment (this is true even in the case of LLR-based approaches [2, 4]). In fact, even in the simple case of a serial system made up of two matchers, selecting these two matchers among N is not easy. In principle,

$$2 \cdot \binom{N}{2} = 2 \cdot \frac{N!}{2 \cdot (N-2)!} = N \cdot (N-1)$$ different

experiments are necessary: this requires time and, consequently, designing serial systems can be much more expensive than adopting parallel fusion. This explains why our model relies on parameters estimated from ROC curves of individual matchers.

The second contributions to the state of the art is as follows. Since ROCs may be affected by estimation errors, in this paper, we show that the model is still useful for designers. This is achieved by modelling estimation errors in order to give to designer a useful support for predicting the acceptability of the system. Analytical findings are supported by preliminary experiments, carried out on the NIST Biometric Score Set 1. They show the reliability of our theoretical analysis.

Paper is organized as follows. Section 2 describes the general model for serial fusion of multiple matchers. Section 3 provides a theoretical analysis of the model when estimation errors occur. Section 4 show experiments for model validation. Conclusions are drawn in Section 5.

## 2. The proposed model

Fig. 1 shows the scheme of the proposed model (see also [3] for the case of two matchers).

In this model, the subject submits to the system the first biometric which is processed and matched against the related template. If the resulting score is more than a predefined upper threshold ($s_{u1}^*$), she/he is accepted as a genuine user. If the score is less than a predefined lower threshold ($s_{l1}^*$), she/he is rejected as an impostor. Otherwise, the system requires a second biometric. The same approach is followed until the final, N-th, matcher is reached. On the basis of it, the subject is finally accepted or rejected.

In the following, we derive the mathematical framework that describes the performance of the whole serial verification system as function of the individual ROC curves of each matcher. In the case of two matchers, it has been shown that several guide-lines for designer can be introduced, when coupling verification rate and verification time [3]. In this paper, we only focus on the verification rate in terms of FAR and FRR.

Let us indicate with $A_i$ the event "the $i$-th system wrongly *A*ccepts the pattern" (false acceptance), and with $R_i$ the event "the $i$-th system wrongly *R*ejects the pattern" (false rejection). We further indicate with $r_{I,i}$ and $r_{G,i}$ the event "the $i$-th matcher *does not classify* the pattern given that it belongs to the impostors class" and "the $i$-th matcher *does not classify* the pattern given that it belongs to the genuine users class", according to the meaning of "rejection option" in pattern classification [6]. In other words, this concept relies on the definition of "uncertainty region", defined by the matching score interval [$s_{li}^*, s_{ui}^*$]. An example is shown in Fig. 2. In this region, the overlapping degree between FAR$_i$ and FRR$_i$ is such that a classification error is very likely to occur. Therefore, no decision is made if match score falls in this interval, and the next biometric in the processing chain is required. Index $i=1,…,N$ indicates the $i$-th matcher of the processing chain.

In order to derive the analytical expression of the whole system's ROC curve, we made the following hypothesis:

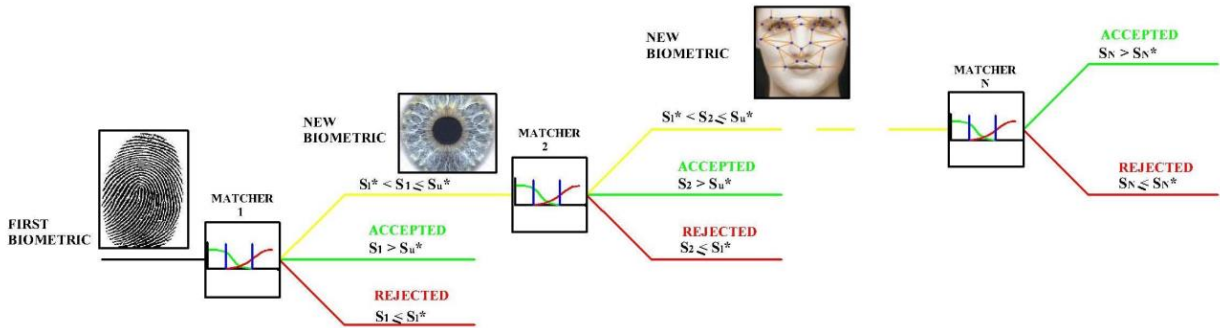

**Fig. 1.** The proposed scheme for serial fusion of multiple matchers. In the scheme, a novel biometry is required if the matcher score between the input pattern and the template falls into the interval between the lower and upper acceptance thresholds.

1. Events $A_i$, $R_i$, $r_{Gi}$ and $r_{Ii}$ are independent each others.

2. Event $r_{Gi}$ and $r_{Ii}$ are such that no genuine users are rejected and no impostors are accepted at $i$-th stage ($i=1,…,N$-1). In other words, lower threshold is correspondent to zeroFRR operational point and upper threshold is correspondent to zeroFAR one. In the following, we indicated such values with *zeroFAR$_i$* and *zeroFRR$_i$* for *i*-th matcher.

Therefore, the probability that the serial system wrongly classifies an impostor (false acceptance), indicated as 'A' event, is as follows:

$$FAR = $$
$$= P(A_1) + P(r_{I1}) \cdot P(A_2) + ... + P(r_{I1}) \cdot ... \cdot P(r_{I(N-1)}) \cdot P(A_N) =$$
$$= FAR_1(s_{I1}^*) \cdot ... \cdot FAR_{N-1}(s_{I(N-1)}^*) \cdot FAR_N(s_N^*)$$

The same hypothesis leads to the following expression of false rejection rate (FRR):

$$FRR =$$
$$= P(R_1) + P(r_{G1}) \cdot P(R_2) + ... + P(r_{G1}) \cdot ... \cdot P(r_{G(N-1)}) \cdot P(R_N) =$$
$$= FRR_1(s_{u1}^*) \cdot ... \cdot FRR_{N-1}(s_{u(N-1)}^*) \cdot FRR_N(s_N^*)$$

In the derivation of these formula, we omitted the intermediate steps which can be easily computed. The final step is to apply constraints desirable for any serial system: (a) the majority of genuine users must be accepted, that is, the operational point related to lower acceptance threshold for matchers *i*=1,…,N-1, must be set to zeroFRR$_i$; (b) all available biometrics must be required for unauthorized users, that is, the operational point related to upper acceptance threshold for matchers *i*=1,…,N-1, must be set to zeroFAR$_i$ [3]. To sum up:

$$FAR = \left[\prod_{i=1}^{N-1} zeroFRR_i\right] FAR(s_N^*) \quad (1)$$

$$FRR = \left[\prod_{i=1}^{N-1} zeroFAR_i\right] FRR(s_N^*) \quad (2)$$

In Eqs. (1-2), we indicated the false acceptance rate and false rejection rate of the second matcher as function of related acceptance threshold $s_2^*$.

The following observations can be made on the basis of Eqs. (1-2):
I. This system meets requirements (a-b) of Section 1 [1].
II. The performance of the whole serial systems is driven by the last matcher.
III. The readability and simplicity of Eqs. (1-2) is mainly due to the independence of decisions among matchers. This may impacts on the performance prediction of the model when such hypothesis is not respected. We investigated by experiments this problem and provided some preliminary results on Section 4.

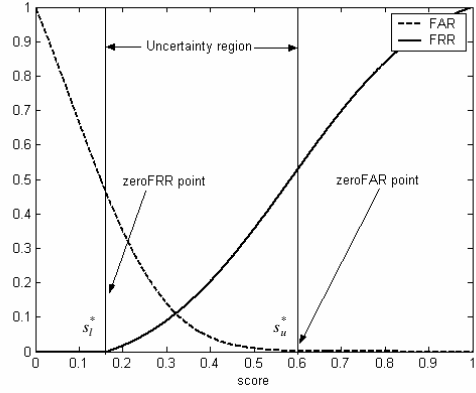

**Fig. 2.** The concept of "uncertainty region" defined by two values of match scores, named "upper" and "lower" acceptance threshold, respectively. If match score falls in this region, the correspondent biometric is considered as not suitable for person verification, and another biometric is required.

IV. In [3], where only the case of *N*=2 is analyzed, the authors observed that, in order to obtain a performance improvement with respect to the "best" individual matcher, it is sufficient to set the processing chain such that the "worst" one is at the first stage. This observation can be generalized. In other words, it is sufficient to put the best matcher at the last stage. No indication is given by Eqs. (1-2) for the previous stages. However, if considering each subset of N-1 matcher as a independent serial system, it is easy to conclude that matcher should be ranked in increasing order of performance (ROC). Obviously, this is not the best processing chain in terms of final ROC, but only the one which assures always an improvement with respect to the best individual matcher, for any operational point. The impact of this conclusion is, however, quite interesting: there is always a configuration of the processing chain such that this serial fusion scheme leads to an improvement of the performance with respect to the best individual matcher, thus justifying, theoretically, the usefulness of combining multiple matchers by serial fusion. It is well-known that a similar conclusion concerning the parallel fusion (e.g. score-level) of multiple matchers cannot be easily reached [1, 7].
V. Another important issue is the estimation of zeroFAR and zeroFRR operational points for each matcher. The impact of estimation errors on the model has not yet been analyzed. This is done here in Section 3, and supported by experiments in Section 4.

## 3. Impact of estimation errors on the proposed model

As pointed out in the previous Section, the proposed model strongly relies on the estimation of zeroFAR and zeroFRR operational points for matchers preceding the last one.

For sake of clarity, we will describe the effect of estimation errors in our model for the case N=2 (only two matchers). The generalization to N matchers, whichever N, is not difficult.

First of all, in order to clearly separate causes and effects, we indicated two kinds of estimation errors:

(1) Error due to bad estimation of the "true" zeroFAR/zeroFRR. In other words, acceptance thresholds are estimated correctly but, for some reasons, e.g. small sample size, zeroFAR/zeroFRR value is not exactly the "true" one. This error will be indicated by the parameter named $\alpha$. Fig. 3 exemplifies such error. We hypothesize that the same error amount is made for zeroFAR and zeroFRR.

(2) Error in the acceptance threshold estimation, even if the estimation of genuine users and impostors probability is correct. This error will be indicated by the parameter $\varepsilon$. Fig. 4 exemplifies such error. We hypothesize that the same error amount is made for zeroFAR and zeroFRR.

In both cases, we also hypothesised that $\alpha$ and $\varepsilon$ displacements are small and, anyway, less than 1.

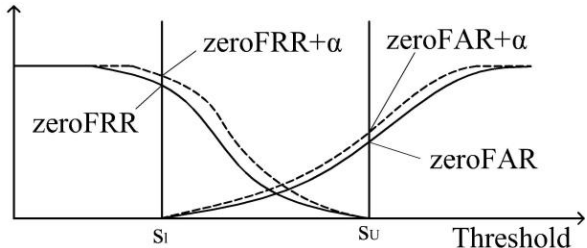

**Fig. 3.** Error due to bad estimation of genuine and impostor distributions, in other words, of FAR and FRR. In this paper, we assume that the same amount $\alpha$ is made for both zeroFAR and zeroFRR. Straight lines are the "true" FAR and FRR, whilst dotted lines are the estimated ones.

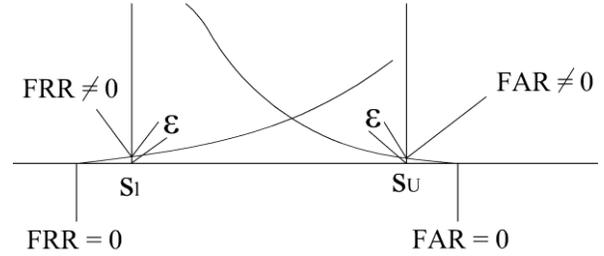

**Fig. 4.** Error due to bad estimation of acceptance thresholds. In this paper, we assume that the same amount $\varepsilon$ is made for both zeroFAR and zeroFRR. In this case, FAR and FRR curves are estimated correctly.

Let $FRR_1(s_l^*)$ and $FAR_1(s_l^*)$ the error rates related to the lower threshold $s_l^*$, and $FRR_1(s_u^*)$ and $FAR_1(s_u^*)$ the error rates related to the upper threshold $s_u^*$, of the first matcher. According to our hypothesis:

$FRR_1(s_l^*) = 0$, $FAR_1(s_l^*) = zeroFRR_1$
$FAR_1(s_u^*) = 0$, $FRR_1(s_u^*) = zeroFAR_1$

Errors reported in items (1)-(2) and Figs. 3-4 are such that:

$$FRR_1(s_l^*) = \varepsilon, \quad FAR_1(s_l^*) = zeroFRR_1 \pm \alpha \quad (3)$$
$$FAR_1(s_u^*) = \varepsilon, \quad FRR_1(s_u^*) = zeroFAR_1 \pm \alpha \quad (4)$$

By introducing Eqs. (3-4) into Eqs. (1-2), and disregarding the terms very near to 0 (typically, second order or superior polynomial on $\alpha$ and $\varepsilon$), we obtain the following expressions for the displacement between expected and "true" ROC of the whole serial system, that we indicate as $\Delta FAR$ and $\Delta FRR$:

$$\Delta FAR = \pm \alpha FAR_2(s_2^*) + \varepsilon \quad (5)$$
$$\Delta FRR = \pm \alpha FRR_2(s_2^*) + \varepsilon \quad (6)$$

The error is linear with $\alpha$, which is the dominant term of Eqs. (5-6), since $\varepsilon$ simply indicates the bias of the error and it is expected to be very low.

The sign of $\alpha$ indicates that zeroFAR/zeroFRR could over- or under-estimated. Both cases can be crucial for a system designer, even if it is well-known that error underestimation can lead to a worst-case scenario.

Eqs. (5-6) have a practical meaning for designers: by setting $\alpha/\varepsilon$ to minimum and maximum values, he can predict, for each operational point of the serial system, the range of the ROC curve for which the performance of the system is still acceptable. This aspect will be shown in Section 4 by experiments.

## 4. Experimental results

### 4.1 Data Set

NIST-Biometric Score Set 1 [8] is made up of scores derived from a fingerprint matcher applied to two different fingers (left and right index), named "LI" and "RI", and two different face matchers named "C" and "G". NIST does not report information about these matchers. The only information is about the number of client, 517, and the number of samples per client, two. Accordingly, we randomly partitioned the data in two parts such that the number of genuine and impostor matching scores per part is 100 and 51,600, as gallery (training) and 417 and 214,656, respectively, as probe set. Training set has been used for estimating the model parameters, namely, lower and upper acceptance thresholds, and operational points of last matcher. No information about the representativeness of adopted templates is available, thus such ROC curves allow to validate the model in both controlled and uncontrolled environments.

### 4.2 Reliability of the model for three matchers

We considered all the four individual systems, thus obtaining different cases.
Table 1 reports the correlation coefficient among matcher pairs. In order to evaluate the reliability of the model under different experimental conditions, we combined serially matchers uncorrelated, and correlated.

**Table 1.** Correlation coefficient among matching scores of pairs of NIST-BSS1 matchers.

|           | Face C | Face G | Finger LI |
|-----------|--------|--------|-----------|
| Face G    | 0.70   | -      | -         |
| Finger LI | -0.12  | -0.13  | -         |
| Finger RI | -0.02  | -0.02  | 0.41      |

Results related to model reliability are shown in Figs. 5-6. Expected ROC according to Eqs. (1-2) have been firstly computed by using estimations on training set. From Fig. 4, a serial system made up of weakly correlated matchers exhibits a good degree of reliability if compared with "real" ROCs computed by performing the estimation by explicitly "fusing" the same data. In other words, if we simply consider the individual ROCs, Eqs. (1-2) allow to derive the whole ROC of the serial system. This ROC is nearly equivalent to the one obtained by submitting each pattern on the whole system, thus "after" connecting matchers, and, accordingly, performing a "real" assessment of the whole system performance. On the other hand, Fig. 6 shows that this prediction is less reliable when considering correlated matchers in the chain. This result points out that the working hypothesis of decision independence is crucial for our framework.

With regard to the observation (IV) of Section 2, we plotted in Fig. 7 two cases, in which the chain with best matcher at the last stage is considered (Fing LI-Face C-Fing RI), and the chain where best and worst matcher have been permutated. Third ROC is related to the best individual matcher (Fing RI). It is worth noting that both ROCs of sequential fusion outperform the best matcher. However, the best combination corresponds to the one where the matcher at the last stage is FingRI, that is, the best individual one. This results is an experimental confirmation of analytical findings in Eqs. (1-2), discussed in Section 2. Although "common" sense should suggest to put the best matcher at the first stage, this experiment clearly shows that this is not the best choice when looking at the overall performance. Vice-versa, this experimental result may appear as counter-intuitive, but analytical findings in Eqs. (1-2) give a simple and clear explanation of it.

As remarked in Section 2, the sequence made up of matchers at increasing level of performance is not the "optimal" one in terms of final ROC. With regard to the problem of setting the most appropriate sequence,, in [3], a condition has been derived. The aim is to help the designer according to the given operational point of the whole system (Eq. 17 in [3]). Currently, this condition is related to the case of two serially combined matchers only. The generalization of this condition will be matter of a future work.

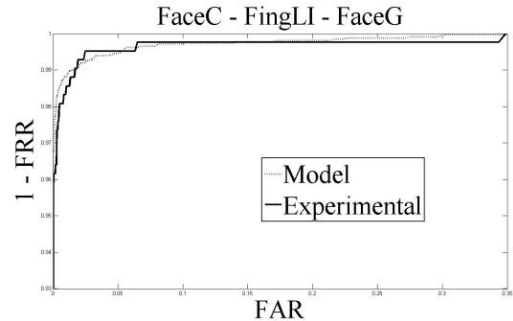

**Fig. 5.** Sequential fusion of FaceC, FingerLI, FaceG. ROC obtained by Eqs (1-2) compared with the one obtained by experiments. It is worth noting that each matcher is weakly correlated with the next one (Table 1).

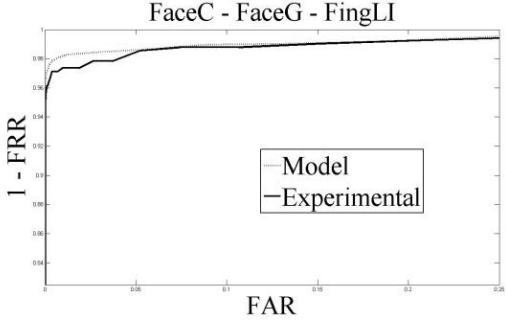

**Fig. 6.** Sequential fusion of FaceC, FaceG, FingerLI. ROC obtained by Eqs (1-2) compared with the one obtained by experiments. In this case, FaceC and FaceG are correlated. Consequently, results are less reliable (Table 1).

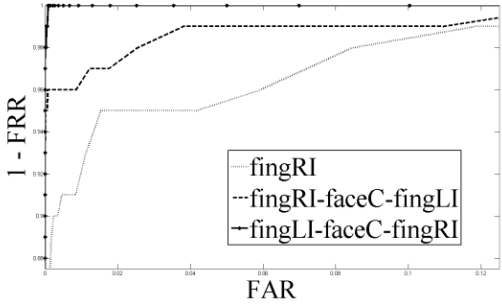

**Fig. 7.** Comparison among sequential fusion and best individual matcher. It is worth noting that, as suggested by Eqs. (1-2), the best sequential fusion is obtained by concatenating matchers by increasing order of performance.

### 4.3 Estimation errors evaluation

In this Section, we will show that,
(a) although hypothesis made in Section 3 are quite simple, they allow to obtain "reliable" ROCs if a good estimation error prediction is provided;
(b) since this estimation is difficult to provide for a system designer, it is possible to estimate "where" the "true" ROC should fall by associating an errors interval dependent on parameters adopted in Eqs. (4-5).

Fig. 8 supports Item (a). We plot ROC curve derived from Eqs. (1-2) (on training set – "Model"), the "corrected" ROC by Eqs. (5-6) ("Model with alpha and epsilon"), where α and ε, according to hypothesis of Section 3, have been estimated on the probe set, and the "true" ROC computed on the probe set itself ("Experimental"). In other words, in this experiment we hypothesised that estimation errors are not known in detail, but computed in terms of α and ε. This is obviously impossible in a real scenario; in other words, an optimal estimation of α and ε is not possible. However, this experiment is aimed to show that these additional parameters can appropriately describe the ROC of the whole serial system when affected by estimation errors. Therefore, Fig. 8 points out that our framework is able to support the case of errors on zeroFAR-zeroFRR operational points. This allows to observe that parameters in Eqs. (5-6) allowed increasing the expressive power of the model of Eqs. (1-2).

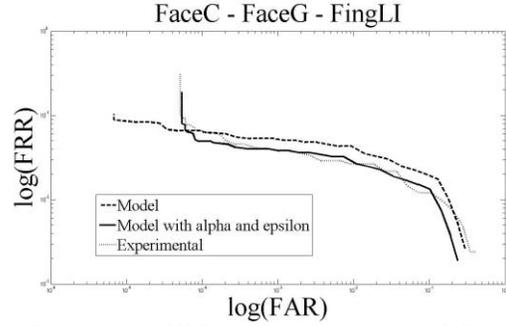

**Fig. 8.** "Model": ROC estimated by Eqs. (1-2) on the training set. "Model with alpha and epsilon": Previous ROC "corrected" by adding α and ε parameters (Eqs. (5-6)) computed on the probe set. "Experimental": ROC estimated on the probe set by experiments.

In order to show Item (b), we plot in Fig. 9 the experimental ROC on the probe set, and two "corrected" ROCs by Eqs. (1-2) on the training set, where α parameter has been set to ±30% of estimated zeroFAR/zeroFRR. In this paper, we considered only α parameter because, from Eqs. (5-6), it appears as the most significant if compared with the bias term ε. This is reasonable, since small sample size problem is always important for biometric matchers, and α quantifies this lacks of information.

The particular α value of Fig. 9 must be only considered as a possible "use-case". In this case, it means that designer accepts an estimation error equal or less than ±30% on estimated zeroFAR/zeroFRR. The less this value, the less the "error" that designer accepts when estimating the system parameters. It is well-known that reliability of a multi-modal biometric system strongly depends on the user population, operational environment, cooperation-level required for each biometric [1]. For example, if it is expected that the real operational environment is nearly equivalent to the training and enrolment conditions of the user population, the designer may reduce the value above. Otherwise, a careful setting is required, in order to have a reasonable prediction of the system behaviour under estimation errors on its parameters. Since many of characteristics which affect a multi-modal biometric system are very difficult to assess quantitatively, it is not possible, at present, to set strong guidelines for assessing these values at best.

However, it is worth noting that our model exhibits the potential of evaluating an errors interval where the "true" ROC may falls in. This interval must be set by considering α and ε values such that ROC achieved are still admissible. Therefore, designer must carefully investigate user population characteristics and intrinsic robustness of individual matchers in front of intra-class variations and environmental conditions which may impact on the whole system robustness.

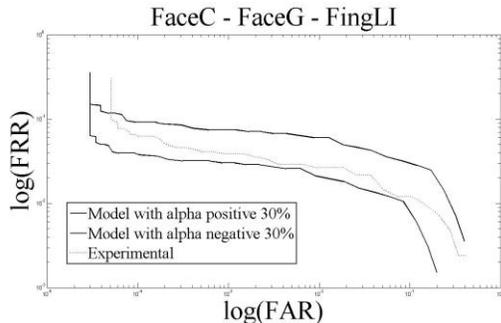

**Fig. 9.** "Experimental": ROC obtained on the probe set. "Model with alpha positive/negative 30%": ROC obtained by Eqs. (1-2), with α set to 30% of estimated zeroFAR/zeroFRR (estimation is done on the training set).

We believe that some of this preliminary results may lead to a better formalization of the estimation error problem when multiple matchers are sequentially combined. Further experiments, on larger multi-modal data sets, could also give more significant insights about the performance and effectiveness of the proposed framework.

## 5. Conclusions

To the best of our knowledge, a few works has dealt with the serial fusion of multiple biometric matchers. If some works proposed the well-known sequential probability ratio test, the authors proposed an original and alternative framework.

In this paper, two contributions to the state-of-the-art have been presented: (1) extended methodology for multi-modal serial fusion for assessing the final system performance on the basis of individual ROC curves, (2) a study on estimation errors on operational points; in other words, we focused on relationships between trained matchers and their generalization ability when unknown samples are submitted.

With regard to item (1), it has been shown that proposed framework allows to reliably predict the whole ROC of the system, and it is particularly effective when low correlated matchers are sequentially connected. With regard to item (2), hypothesis and theoretical results achieved allows the designer to set an appropriate errors interval, where the whole ROC may be considered still acceptable during system's operations.

Above points are supported by experiments on well-known benchmark data set.

Further works will rely on performing experiments on larger data sets, and the full theoretical and experimental assessment of proposed framework robustness and reliability, in terms of verification rate and time. Comparison with alternative approaches at the state-of-the-art will be also matter of future works.